\documentclass[sigconf]{acmart}

\AtBeginDocument{%
  }

\setcopyright{none}
\renewcommand\footnotetextcopyrightpermission[1]{}

\begin{document}
\pagestyle{empty}

\title{SPICE: Synergy and Partial Information Based Curriculum Evolution}

\author{Ankush Pratap Singh}
\affiliation{%
  \department{Computer Science}
  \institution{New York Institute of Technology}
  \city{New York}
  \state{New York}
  \country{USA}
}
\email{asing213@nyit.edu}

\author{Houwei Cao}
\affiliation{%
  \department{Computer Science}
  \institution{New York Institute of Technology}
  \city{New York}
  \state{New York}
  \country{USA}
}
\email{hcao02@nyit.edu}

\author{Yong Liu}
\affiliation{%
  \department{Electrical and Computer Engineering}
  \institution{New York University}
  \city{Brooklyn}
  \state{New York}
  \country{USA}
}
\email{yongliu@nyu.edu}

\renewcommand{\shortauthors}{Singh et al.}

\begin{abstract}
Multimodal learning exploits complementary information across heterogeneous modalities. The informativeness of each modality can vary widely across samples and training stages. Existing multimodal curriculum learning strategies often assume that the relative complexity of samples remains unchanged throughout training and therefore cannot adapt to model evolution. We propose \textbf{SPICE} (\textbf{S}ynergy and \textbf{P}artial \textbf{I}nformation based \textbf{C}urriculum \textbf{E}volution), a novel progressive curriculum framework for multimodal interaction learning. Guided by \textbf{Partial Information Decomposition (PID)} theory, our approach decomposes multimodal interactions into \textit{redundant}, \textit{unique}, and \textit{synergistic} information components, enabling an interpretable and dynamic characterization of sample complexity. Building on this decomposition, we design a progressive curriculum that evolves throughout training, allowing the model to transition from learning shared cross-modal cues to modality-specific patterns and, finally, to complex synergistic interactions. Adapting to model evolution, sample ordering is refined in real-time using PID information estimates derived from unimodal and multimodal predictions. Experiments across multiple multimodal benchmarks demonstrate consistent improvements over conventional training and state-of-the-art baselines, highlighting the effectiveness of PID information decomposition and adaptive sample ordering for multimodal curriculum learning.
\end{abstract}




\maketitle

\section{Introduction}

Multimodal learning has become a fundamental paradigm for intelligent interactive systems, enabling models to jointly reason across heterogeneous modalities such as audio, vision, text, and motion~\cite{baltruvsaitis2018multimodal, yuan2025}. By leveraging complementary information across modalities, multimodal models have achieved strong performance across a wide range of tasks, including affective computing~\cite{affectivecomputingpicard,affectivecomputingpei}, audiovisual understanding~\cite{audiovideounderstandingalayrac2020self,audiovideounderstandingwei2022learning}, human-computer interaction~\cite{hcisharma,hciazofeifa2022systematic}, and multimodal reasoning~\cite{multimodalreaosninglu2022learn,multimodalreasoningwang2024exploring}. A key challenge in these systems is effectively learning \emph{cross-modal interactions}, as different modalities often contribute unequally, the relative informativeness of modalities varies across training samples, and model components specific for different modalities often converge at different rates throughout training~\cite{wang2019WhatMT,peng2022ogm}.

\begin{figure}[t]
  \centering
  \includegraphics[width=0.92\linewidth]{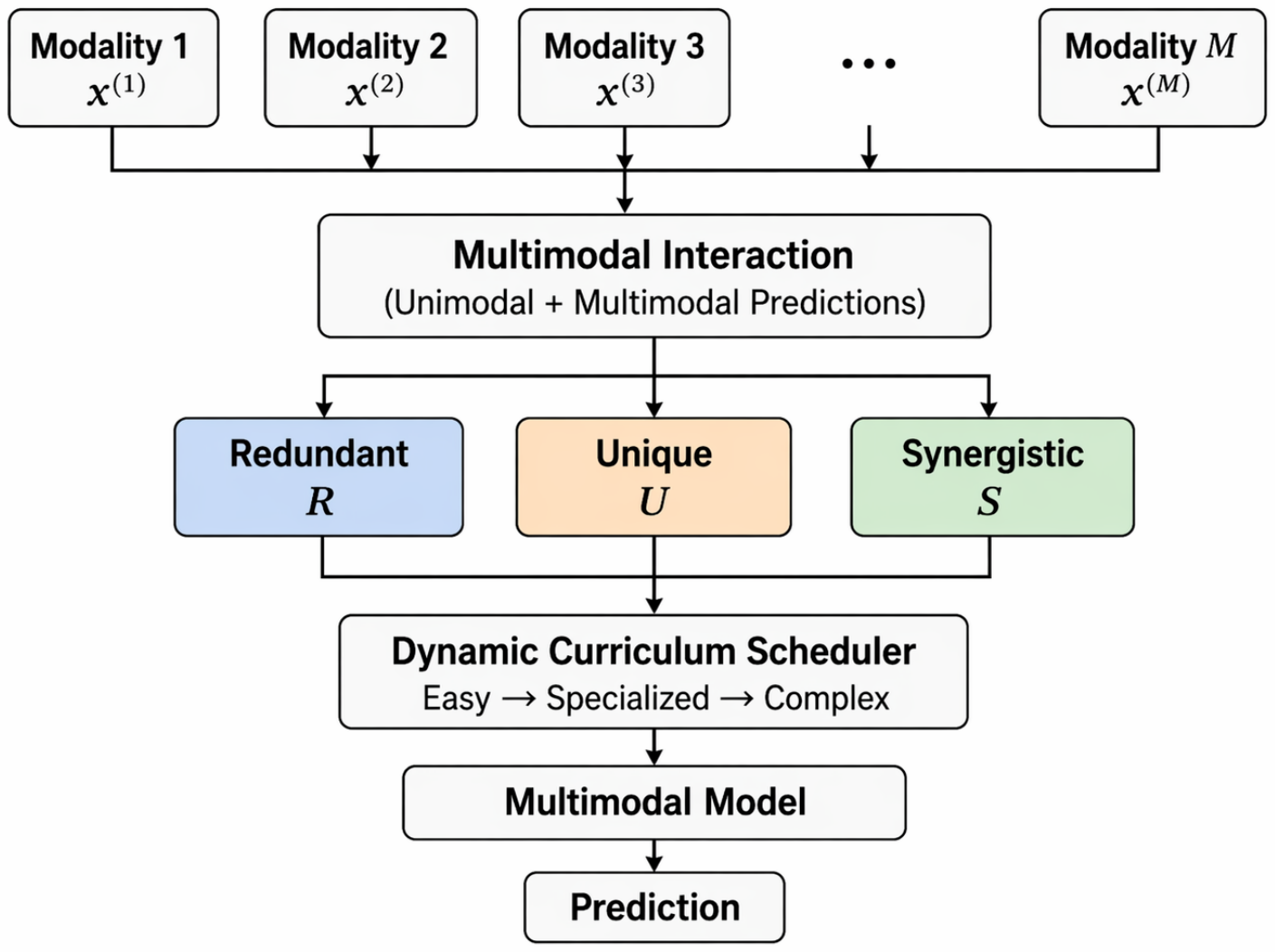}
  \caption{SPICE Workflow: PID Information Estimation $\Rightarrow$ Adaptive Sample Ordering $\Rightarrow$ Dynamic Curriculum}
  \Description{A flowchart illustrating the decomposition of multimodal interactions into R, U, and S, and dynamically scheduling the curriculum for multimodal prediction.}
  \label{fig:teaser-fig}
\end{figure}

\begin{figure*}
  \centering
  \includegraphics[width=0.98\linewidth]{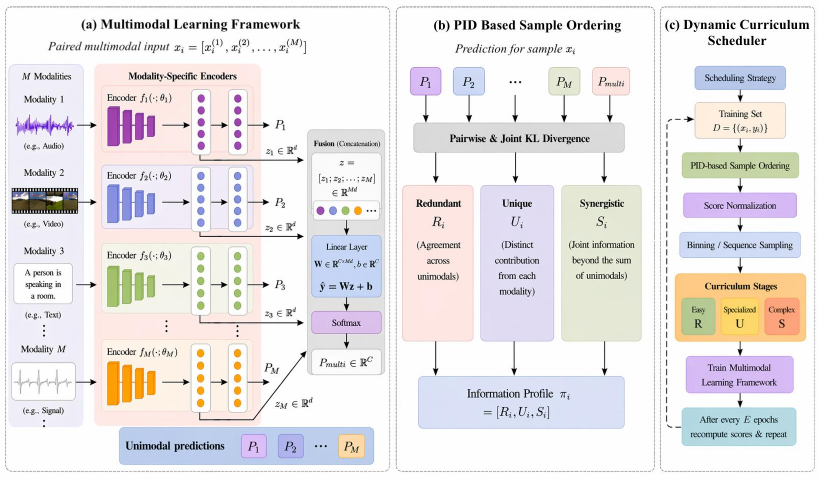}
  \caption{Illustration of SPICE: (a). Multi-modal learning framework, (b). PID-Based Sample Ordering, (c). Dynamic Curriculum Scheduler.}
  \Description{Illustration of SPICE in a block diagram and flowchart}
  \label{fig:block-diagram}
\end{figure*}

Existing multimodal training strategies typically assume that the relative difficulty of samples and modalities remains fixed throughout model weights optimization. Under this view, samples are either treated uniformly via random shuffling or assigned static difficulty scores for curriculum-based training~\cite{bengio2009curriculum,hacohen2019power,ankushchuckle,soviany2022curriculum}, which is a training paradigm in which samples are presented in a meaningful order, typically progressing from easier to more difficult examples, to improve optimization and generalization. However, this assumption is fundamentally limiting. In practice, sample difficulty is inherently \emph{dynamic}: samples that are difficult early in training may become easier as the model and representations mature~\cite{Wang_2019_ICCV}. For multimodal interaction learning, a sample's complexity depends not only on the quality of each modality but also on how the modalities interact. For example, some samples may contain strongly aligned cross-modal cues that are straightforward for the model to learn. In contrast, other samples might require the model to identify subtle signals specific to each modality or to recognize higher-order synergistic dependencies that become informative only when signals from multiple modalities are  combined~\cite{barrett2015exploration}. Higher-order cross-modal interactions may emerge only after strong modality-specific features are learned. Ignoring this evolving nature of multimodal training can lead to suboptimal training dynamics and poor modeling of complex dependencies between modalities.

To address this, we propose \textbf{SPICE} (Figure \ref{fig:teaser-fig}), a novel \textbf{progressive curriculum framework} that dynamically assesses the difficulty of multimodal samples throughout training. Guided by the \textbf{Partial Information Decomposition (PID)} theory~\cite{Williams2010NonnegativeDO, Bertschinger2013QuantifyingUI}, our approach captures multimodal interactions through three interpretable components: \emph{redundant}, \emph{unique}, and \emph{synergistic} information. At any training stage, we quantify a sample's information distribution across the three PID components based on agreement/disagreement among the current model’s unimodal and multimodal predictions, as well as their alignment with its ground-truth label. A learning curriculum is constructed by strategically ordering samples according to their PID information distributions, guiding the model from learning shared, easy, low-level cross-modal representations to recognizing modality-specific representations, and finally to understanding complex synergistic interactions. Unlike conventional curricula, our framework continuously updates the sample ordering based on real-time unimodal and multimodal predictions, adapting to the model’s evolving internal learning state~\cite{jiang2021selfpaced, graves2017automated}. Figure~\ref{fig:block-diagram} presents an overview of the proposed framework, including (a) the multimodal learning architecture, (b) PID-based sample ordering, and (c) dynamic curriculum scheduler.

We evaluate the proposed approach across multiple multimodal benchmarks and demonstrate consistent improvements over the traditional training strategies and state-of-the-art baselines. Our work demonstrates that an information-theoretic decomposition provides a robust framework for designing curricula in multimodal learning. Our results also emphasize the importance of dynamically assessing sample difficulty to make the curriculum adaptive to the training state. The main contributions are summarized as follows:
\begin{itemize}
 \item We propose a novel multimodal curriculum learning framework guided by PID information theory, which orders samples based on their PID information distributions. This approach helps the model progressively learn redundant, unique, and synergistic multimodal interactions.
 
 \item We emphasize that sample difficulty in multimodal learning is inherently dynamic and evolves with the model's training state. Using real-time PID information estimates from unimodal and multimodal predictions, our curriculum adapts sample ordering to the model's evolution.   
 
 \item We demonstrate consistent improvements across multiple multimodal datasets and tasks compared to conventional and state-of-the-art multimodal training baselines.
\end{itemize}

\section{Related Work}

\subsection{Multimodal Interaction Learning}

Multimodal learning aims to jointly model heterogeneous data sources, such as audio, vision, text, and motion, to capture complementary information for downstream tasks. Recent advances in multimodal learning have shown strong performance in areas such as audiovisual understanding, sentiment analysis, emotion recognition, and human-computer interaction. A significant challenge for these systems is learning \emph{cross-modal interactions} effectively.  

Existing approaches primarily focus on improving fusion architectures, cross-modal attention mechanisms, and alignment strategies to better understand interactions among modalities ~\cite{perez2018film, nie2021multimodal,han2023trusted}. While these methods enhance representation learning, they often overlook how the \emph{difficulty of learning multimodal interactions evolves during training}. Our work explicitly models interaction complexity as a dynamic quantity that changes with the model's learning state.

\subsection{Balanced and Curriculum-based Multimodal Learning}

Curriculum learning, introduced by Bengio et al.~\cite{bengio2009curriculum}, highlights the benefits of presenting training samples in order of increasing difficulty. Subsequent research extended this idea to self-paced learning~\cite{jiang2021selfpaced}, in which models select informative samples during training, and to automated curriculum learning~\cite{graves2017automated}, which adapts curricula based on the training state. These approaches demonstrate that sample difficulty can change and should be reassessed throughout optimization. However, in multimodal learning, dynamic curriculum strategies are less explored, with research mainly focusing on balancing optimization across modalities to prevent stronger ones from overshadowing weaker ones.

A growing body of research has addressed the issue of \emph{modality imbalance}, in which dominant modalities can suppress weaker ones during joint optimization. Prior methods have tackled this challenge from various perspectives. \emph{Gradient modulation} techniques rebalance optimization dynamics by adjusting gradient contributions across modalities, as explored in OGM~\cite{peng2022ogm} and AGM~\cite{li2023agm}. Additionally, \emph{prototype- and representation-based balancing} methods, such as PMR~\cite{fan2023pmr}, MLA~\cite{zhang2024mla}, and AMSS~\cite{yang2025amss}, enhance modality cooperation through structured feature adaptation and representation alignment. More recently, \emph{sample-level and optimization-aware strategies} such as ReconBoost~\cite{hua2024reconboost}, MMPareto~\cite{wei2024mmpareto}, and SMV~\cite{wei2024smv} have sought to improve multimodal robustness through sample valuation, Pareto-based optimization, and modality reconciliation.

Balance-aware sequence sampling methods, such as BSS~\cite{guan2025balanceaware}, have also shown that the order in which samples are presented can significantly enhance multimodal learning by transitioning from balanced to imbalanced samples. However, existing curriculum strategies often rely on difficulty measures such as loss and prediction similarity, rather than on agreement between the modalities and alignment with the ground truth. In contrast, our proposed SPICE framework dynamically adjusts sample ordering based on alignment and agreement of current unimodal and multimodal predictions, enabling progressive learning that evolves from redundant to unique and synergistic interactions as the model matures.

\subsection{Partial Information Decomposition (PID)}

Information-theoretic frameworks have long been principled tools for analyzing multimodal interactions. In particular, \textbf{Partial Information Decomposition (PID)} ~\cite{Williams2010NonnegativeDO, Bertschinger2013QuantifyingUI} decomposes multimodal information into \emph{redundant}, \emph{unique}, and \emph{synergistic} components. In an example of predicting $Y$ using 
 two modalities $X_1$ and $X_2$:   
\[
I(X_1,X_2;Y)= R(Y)+U_{X_1}(Y)+U_{X_2}(Y)+S(Y),
\]
 where $R(Y)$ is redundant information about $Y$ present in \emph{both} modalities, $U_{X_i}(Y)$ is information unique to $X_i$, and synergistic information $S(Y)$ arises \emph{only} under joint observation of $X_1$ and $X_2$.  PID extends naturally to more than two modalities \cite{griffith2014quantifying}.

PID provides an interpretable characterization of how different modalities contribute individually and jointly to prediction. Recent studies have explored PID for multimodal representation analysis and interpretability \cite{ma2025pid}. However, these approaches primarily focus on post-hoc analysis rather than training dynamics. To the best of our knowledge, our work is the first to leverage PID as a \emph{dynamic curriculum signal} for multimodal interaction learning.

\section{Methodology}

In this section, we present \textbf{SPICE}, a PID-guided curriculum framework for multimodal interaction learning. The proposed method is motivated by the observation that multimodal sample difficulty is fundamentally governed by the distribution of information across modalities and its evolution during training. Some samples contain strong shared information across multiple modalities and are easier to learn, while others require modality-specific reasoning or higher-order cross-modal interactions. To explicitly model this behavior, we decompose multimodal information into three components: \textbf{redundancy}, \textbf{unique information}, and \textbf{synergy}. Based on these quantities, we design a progressive curriculum learning strategy that dynamically updates sample ordering using unimodal and multimodal predictions.

\subsection{Problem Formulation}

Let the multimodal training dataset be defined as
\begin{equation}
\mathcal{D} =
\left\{
\left(
\{x_i^{(m)}\}_{m=1}^{M},
y_i
\right)
\right\}_{i=1}^{N},
\end{equation}
where \(N\) is the number of samples, \(M\) is the number of modalities, \(x_i^{(m)}\) denotes the \(m\)-th modality of the \(i\)-th sample,  and \(y_i \in \{1,\dots,C\}\) is its class label. For each modality, we use a modality-specific encoder:
\begin{equation}
z_i^{(m)} = f_m(x_i^{(m)}),
\quad m=1,\dots,M, 
\end{equation}
where \(f_m(\cdot)\) denotes the encoder corresponding to modality \(m\). The fused multimodal representation is obtained as
\begin{equation}
z_i^{multi} =
g\left(
z_i^{(1)}, z_i^{(2)}, \dots, z_i^{(M)}
\right),
\end{equation}
where \(g(\cdot)\) denotes the multimodal fusion function. The unimodal prediction distributions are: 
\begin{equation}
p_i^{(m)} =
\text{softmax}(W_m z_i^{(m)}), \quad m=1,\dots,M, 
\end{equation}
while the fused multimodal prediction is: 
\begin{equation}
p_i^{multi} =
\text{softmax}(W_{multi} z_i^{multi}).
\end{equation}
The confidence scores for unimodal and multi-modal predictions with respect to the ground-truth label $y_i$ are defined as: 
\begin{equation}
c_i^{(m)} = p_i^{(m)}(y_i),\quad c_i^{multi} = p_i^{multi}(y_i).
\end{equation}

\subsection{PID-inspired Sample Ordering}

Motivated by PID theory, for sample $i$, the mutual information between its multimodal data $\{x_i^{(m)}\}_{m=1}^{M}$ and the target label $y_i$ can be decomposed into information \textit{redundant} across all modalities, information \textit{unique} to a single modality, and \textit{synergistic} information that only presents through the interactions of all modalities. The exact PID estimation for high-dimensional continuous multimodal representations is computationally intractable and remains an active research problem. Instead, we construct measures that capture the characteristic behavior of redundant, unique, and synergy using the current unimodal and multimodal prediction distributions. These scores provide an interpretable approximation of the PID information while remaining computationally efficient and suitable for continual updates throughout the training.

For multimodal training, samples carrying mostly redundant information are the ``easiest", and can be used to bootstrap encoders for all modalities; samples with unique information in a dominating expressive modality are specialized to train the encoder for that modality; samples with strong synergistic information are ideal candidates to train multimodal fusion. This naturally leads to a progressive multimodal curriculum with ordered samples: 
\begin{equation}
\label{eq:stages}
\textit{redundant (easy)} \rightarrow \textit{unique (specialized)} \rightarrow \textit{synergistic (complex)}
\end{equation}
To quantify the distribution of PID information, one can rely on \textit{static exogenous inputs}, e.g., unimodal and multimodal human perceptions of emotion clips in CREMA-D~\cite{cao2014cremad}. If unimodal and multimodal human perceived emotion labels match with ground-truth labels (the intended emotion), the clip is modality-redundant; if only audio or visual perception matches with the intended, the clip is modality-unique; if neither audio nor visual perception matches with the ground-truth, but audio-visual perception does, the clip carries mostly synergistic information. {\it However, many multimodal datasets lack per-modality labels. In addition, as discussed earlier, static sample ordering cannot adapt to model training evolution.}

\subsection{Dynamic Multimodal PID Curriculum}
To develop a dynamic curriculum for any multimodal datasets, we propose to estimate PID information distribution for sample $i$ using \textit{endogenous model outputs}, namely unimodal and multimodal predictions 
$\{\{p_i^{(m)}\}_{i=1}^M, p_i^{multi}\}$. We consider two key factors:  \textbf{agreement/disagreement across unimodal and multimodal predictions}, and \textbf{their alignment with the ground-truth label}. This dual perspective allows SPICE to distinguish easy shared cues from modality-specific and synergistic interactions. 
Throughout training, we periodically refresh the PID information distribution estimates, allowing the curriculum to adaptively reorder samples as the model matures.    

\subsubsection{Redundancy Score}
For a sample that carries mostly redundant information, all modalities independently provide strong, mutually consistent predictions that align with the ground truth. For unimodal prediction in modality \(m\), we first compute its divergence from the average prediction in the remaining modalities:
\begin{equation}
\label{eq:divergence}
D_i^{(m)}
=
D_{KL}
\left(
p_i^{(m)}
\parallel
\frac{1}{M-1}
\sum_{n\neq m}p_i^{(n)}
\right), 
\end{equation}
where $D_{KL}(\cdot)$ is the  directional Kullback-Leibler (KL) divergence. We then compute the average cross-modal divergence as
\begin{equation}
\bar D_i
=
\frac{1}{M}
\sum_{m=1}^{M}
D_i^{(m)}.
\end{equation}
The redundancy score for sample $i$ is defined as
\begin{equation}
R_i
\triangleq
\underbrace{
\left(
\prod_{m=1}^{M}
c_i^{(m)}
\right)}_{\text{alignment}}
\cdot
\underbrace{
\exp(-\bar D_i)}_{\text{agreement}}, 
\end{equation}
where the alignment term ensures that all modalities confidently assign high values to the ground-truth label, while the agreement term penalizes a mismatch between unimodal predictions. A high redundancy score indicates that the modalities provide strong, consistent predictions aligned with the true label.

\subsubsection{Unique Information Score}
For a sample dominated by unique information, one modality provides strong discriminative evidence while the remaining modalities are relatively uncertain. For each modality \(m\), we define its unique contribution as
\begin{equation}
U_i^{(m)}
\triangleq
\underbrace{
c_i^{(m)}
\left(
\prod_{n\neq m}
(1-c_i^{(n)})
\right)}_{\text{alignment}}
\cdot
\underbrace{
\left(
1-\exp(-D_i^{(m)})
\right)}_{\text{agreement}},
\end{equation}
where \(D_i^{(m)}\) is the KL-divergence of modality \(m\) from the average prediction defined in (\ref{eq:divergence}). The first term captures samples in which the $m$-th modality is highly confident about the ground truth, whereas the other modalities exhibit lower confidence. The second term ensures that modality $m$ provides a distinct prediction from the other modalities rather than simply varying in confidence magnitude. A higher unique information score indicates a greater degree of modality-specific learning.

The total unique score is obtained by summing up the unique scores in all modalities:
\begin{equation}
U_i
\triangleq
\sum_{m=1}^{M}
U_i^{(m)}.
\end{equation}

\subsubsection{Synergy Score}

Synergy captures higher-order information that emerges only after multimodal fusion and is not available in any individual modality. We first calculate the divergence of multimodal prediction from unimodal predictions: 
\begin{equation}
D_i^{syn}
=
D_{KL}
\left(
p_i^{multi}
\parallel
\frac{1}{M}
\sum_{m=1}^{M}
p_i^{(m)}
\right).
\end{equation}
The synergy score is computed as
\begin{equation}
S_i
\triangleq
\underbrace{
\left(
c_i^{multi}
-
\max_m c_i^{(m)}
\right)}_{\text{alignment}}
\cdot
\underbrace{
D_i^{syn}}_{\text{agreement}}.
\end{equation}
The alignment term assesses how much the fused prediction improves over the strongest individual modality. Meanwhile, the agreement term measures how much the multimodal prediction differs from the average of all unimodal predictions. A higher synergy score indicates larger information gain from cross-modal interactions beyond unimodal information.

\subsubsection{Score Normalization}

Since \(R_i\), \(U_i\), and \(S_i\) may lie on different scales, we normalize each component using min-max normalization.
\begin{equation}
\hat{C}_i =
\frac{C_i-C_{min}}{C_{max}-C_{min}},
\quad
C \in \{R, U, S\}.
\end{equation}
This ensures $\hat{R}_i,\hat{U}_i,\hat{S}_i \in [0,1]$.

\subsection{Sample Allocation Strategies}
As described in (\ref{eq:stages}), samples are progressively introduced to curriculum learning based on their PID information distributions. We investigate two sample allocation strategies. 

\subsubsection{SPICE-S: Stage-wise Curriculum with Sample Binning}
In the stage-wise curriculum, we first divide samples into three bins: redundant $\mathcal{D}_R$, unique $\mathcal{D}_U$, and synergistic $\mathcal{D}_S$   based on their dominant PID scores, i.e.,  $\arg\max(\hat{R}_i,\hat{U}_i,\hat{S}_i)$. 
Training begins with only samples from the Redundant bin. Samples from the Unique and Synergy bins are added sequentially in the following stages, facilitating gradual learning while reducing the risk of catastrophic forgetting.
\begin{equation}
(\mathcal{D}_R)
\rightarrow
(\mathcal{D}_R \cup \mathcal{D}_U)
\rightarrow
(\mathcal{D}_R \cup \mathcal{D}_U \cup \mathcal{D}_S).
\end{equation}
One motivation of SPICE-S is to lower the computation cost associated with gradient updates because it is trained using a subset of the dataset rather than the entire dataset in the first two stages. If one gradient update is performed per batch, SPICE-S requires fewer gradient updates to complete the same number of training epochs. A more detailed discussion on this can be found in \ref{sec:compairng-SPICES-SPICEE}.
To adapt to model evolution, the PID scores are recalculated every $k$ epochs. Samples are redistributed to bins using updated PID scores. 

\subsubsection{SPICE-E: Entire Dataset Curriculum with Sample Ordering}
 In SPICE-E, the entire dataset is used throughout training. Instead of assigning samples to bins based on their dominant PID scores, we use different PID scores to determine the sample orders in different training stages. We first calculate the relative contribution of each PID component for a given sample as  
  $P_R(i)=\hat{R}_i /\hat T_i$, $P_U(i)=\hat{U}_i/\hat T_i$, $P_S(i)=\hat{S}_i /\hat T_i$, where $\hat T_i=\hat{R}_i+\hat{U}_i+\hat{S}_i$. 
  
During each epoch of all training stages (Redundant, Unique, or Synergistic), all samples are sequenced based on stage-specific sampling probabilities derived from their PID scores. A sample drawn into an earlier batch will be excluded from the subsequent batches in the same epoch. Specifically, in the Redundant stage, higher $P_R$ scores indicate easier samples, so the sampling probabilities are proportional to \( P_R \), prioritizing more redundant instances in the earlier batches. In contrast, in the Unique and Synergistic stages, higher $P_U$ and $P_S$ scores signify increased training difficulty. To facilitate an easy-to-hard learning progression, we use \( 1 - P_U \) and \( 1 - P_S \) as the sampling probabilities in the Unique and Synergistic stages, respectively, prioritizing simpler samples to the earlier batches. The PID scores are recalculated every $k$ epochs, making the sample ordering adaptive to model evolution.  

\subsection{Loss Function}
The total loss is defined as
\begin{equation}
\mathcal{L}
=
\mathcal{L}_{multi}
+
\sum_{m=1}^{M}\mathcal{L}_m, 
\end{equation}
where $\mathcal{L}_{multi}=CE(p^{multi},y)$ and $\mathcal{L}_m=CE(p^{(m)},y)$ are the cross-entropy losses for multimodal and unimodal predictions.  

\subsection{Model Inference}

During inference, we follow the standard multimodal late-fusion approach, as established in BSS~\cite{guan2025balanceaware}, MLA~\cite{zhang2024mla}, and DI-MML~\cite{DI-MML}. The final prediction aggregates unimodal and multimodal logits. The co-trained unimodal heads additionally provide a free ensemble that helps mitigate strong-modality dominance during prediction.

\begin{equation}
 p_{\text{logits}}^{final}
=
 p_{\text{logits}}^{multi}
+
\sum_{m=1}^{M}
 p_{\text{logits}}^{(m)}.
\end{equation}
The predicted class label is
\begin{equation}
\hat{y}
=
\arg\max( p_{\text{logits}}^{final}).
\end{equation}

\section{Experiments}

In this section, we present our evaluation of the proposed \textbf{SPICE} framework across various multimodal benchmarks, including audiovisual emotion recognition, audiovisual event understanding, and trimodal gesture recognition. Our experiments are designed to address three key questions:
(i) \textit{Does dynamically modeling modality interactions and sample difficulty enhance multimodal learning?}
(ii) \textit{How does SPICE compare to existing state-of-the-art multimodal interaction baselines?}
(iii) \textit{What roles do the individual PID components play in the construction of multimodal curriculum?}

\subsection{Datasets}

To evaluate the general applicability of the proposed SPICE framework across diverse multimodal interaction scenarios, we conducted experiments using four widely recognized benchmark datasets. These cover different areas, such as emotion recognition, audiovisual event understanding, and gesture recognition, as well as datasets of different scales.

\textbf{CREMA-D}~\cite{cao2014cremad} is a multimodal emotion recognition dataset comprising audiovisual clips from 91 actors who perform emotional expressions across six categories: anger, disgust, fear, happiness, neutrality, and sadness. The dataset includes a total of 7,442 video clips, with 6,698 samples designated for training and 744 for testing.

\textbf{Kinetics-Sounds}~\cite{willkay2017} is an audiovisual action recognition benchmark containing short video clips paired with corresponding audio signals. It comprises 31 distinct action classes, with 15,000 training samples and 1,900 testing samples.

\textbf{NVGesture}~\cite{pavlo2016} is a multimodal gesture recognition dataset that encompasses RGB, optical flow, and depth modalities. It consists of 25 dynamic hand gesture classes captured from multiple viewpoints and sensors, making it particularly suitable for evaluating trimodal interaction learning. The dataset includes 1,050 samples for training and 482 samples for testing.

\textbf{VGGSound}~\cite{chen2020vggsound} is a large-scale audiovisual benchmark that features video clips collected from diverse real-world environments. It includes 310 categories and presents a challenging setting for large-scale multimodal interaction learning due to significant variations in visual and acoustic contexts. The dataset comprises 182,608 videos for training, along with 15,337 videos for testing.

\begin{table}[t]
\centering
\caption{Implementation details across datasets.}
\label{tab:impl_details}
\small
\begin{tabular}{lcccc}
\toprule
\textbf{Setting} & \textbf{CREMA-D} & \textbf{KS} & \textbf{NVG} & \textbf{VGGS} \\
\midrule
Modalities     & 2        & 2        & 3        & 2 \\
Backbone       & ResNet18 & ResNet18 & ResNet18 & ResNet18 \\
Optimizer      & SGD      & SGD      & SGD      & SGD \\
Learning Rate  & 0.01     & 0.1      & 0.01     & 0.01 \\
Momentum       & 0.9      & 0.9      & 0.9      & 0.9 \\
Weight Decay   & $10^{-4}$ & $10^{-4}$ & $10^{-4}$ & $10^{-4}$ \\
Scheduler      & Cosine & Step & Step & Step \\
\bottomrule
\end{tabular}
\end{table}

\subsection{Implementation Details}

For our work, we follow the standard ResNet18 architectures as the backbone to isolate the impact of curriculum sampling from architectural complexities. The detailed implementation settings are summarized in Table~\ref{tab:impl_details}. 

For bimodal datasets such as CREMA-D, Kinetics-Sounds, and VGGSound, we employ separate encoders and linear layers for modality representations and predictions. These representations are then processed by a fusion module for multimodal representation, followed by a linear prediction head.

For NVGesture, we use three independent encoders and prediction layers for the RGB, optical flow, and depth branches. The representations are subsequently combined through multimodal fusion and passed to a linear layer for prediction.

All models are trained for 150 epochs. For PID curricula, each training stage consists of 50 epochs. An initial warm-up stage of 30 epochs is introduced to stabilize representations before dynamic curriculum learning begins. This approach prevents the curriculum from being biased towards a particular PID component due to unstable early predictions and encourages faster convergence. 

Following our methodology, PID scores are recalculated every $k=5$ epochs using the latest model predictions. These dynamically updated PID scores are then used to allocate and order samples in new PID curricula.

\begin{table*}
\caption{Performance comparison on CREMA-D, Kinetics-Sounds, and NVGesture datasets. Best results are shown in \textbf{bold}.}
\label{tab:main_results}
\small
\begin{tabular}{lcccccc}
\toprule
\textbf{Method} 
& \multicolumn{2}{c}{\textbf{CREMA-D}} 
& \multicolumn{2}{c}{\textbf{Kinetics-Sounds}} 
& \multicolumn{2}{c}{\textbf{NVGesture}} \\
\cmidrule(lr){2-3} \cmidrule(lr){4-5} \cmidrule(lr){6-7}
& ACC (\%) & mAP (\%) & ACC (\%) & mAP (\%) & ACC (\%) & F1 (\%) \\

\midrule
Audio / RGB & 63.17 & 68.61 & 54.12 & 56.69 & 78.22 & 78.33 \\
Video / OF & 45.83 & 58.79 & 55.62 & 58.37 & 78.63 & 78.65 \\
Depth & -- & -- & -- & -- & 81.54 & 81.83 \\
\midrule
Concat & 63.31 & 68.31 & 64.55 & 71.31 & 81.33 & 81.47 \\
Affine & 66.26 & 71.93 & 64.24 & 69.31 & 82.78 & 82.81 \\
Channel & 66.13 & 71.75 & 63.51 & 68.66 & 81.54 & 81.57 \\
ML-LSTM & 62.94 & 64.73 & 63.84 & 69.02 & 83.20 & 83.30 \\
Sum & 63.44 & 69.08 & 64.97 & 71.03 & 82.99 & 83.05 \\
Weight & 66.53 & 73.26 & 65.33 & 71.33 & 83.42 & 83.57 \\
ETMC & 65.86 & 71.34 & 65.67 & 71.19 & 83.61 & 83.69 \\
\midrule
MSES & 61.56 & 68.83 & 64.71 & 70.63 & 81.12 & 81.47 \\
OGR-GB & 64.65 & 84.54 & 67.10 & 71.39 & 82.99 & 83.05 \\
DOMFN & 67.34 & 85.72 & 66.25 & 72.44 & -- & -- \\
OGM & 66.94 & 71.73 & 66.06 & 71.44 & -- & -- \\
MSLR & 65.46 & 71.38 & 65.91 & 71.96 & 82.86 & 82.92 \\
AGM & 67.07 & 73.58 & 66.02 & 72.52 & 82.78 & 82.82 \\
PMR & 66.59 & 70.30 & 66.56 & 71.93 & -- & -- \\
ReconBoost & 74.84 & 81.24 & 70.85 & 74.24 & 84.13 & 86.32 \\
MMPareto & 74.87 & 85.35 & 70.00 & 78.50 & 83.82 & 84.24 \\
SMV & 78.72 & 84.17 & 69.00 & 74.26 & 83.52 & 83.41 \\
MLA & 79.43 & 85.72 & 70.04 & 74.13 & 83.40 & 83.72 \\
AMSS & 70.30 & 76.14 & 72.25 & 79.13 & 84.64 & 84.94 \\
BSS-H & 80.78 & 87.86 & 72.67 & 78.61 & 85.06 & 85.15 \\
BSS-L & 82.80 & 88.61 & 73.95 & 79.43 & 86.72 & 87.04 \\
\midrule
SPICE-S & 80.91 & 86.76 & 72.41 & 78.07 & 86.51 & 86.77 \\
SPICE-E & \textbf{83.06} & \textbf{89.07} & \textbf{73.99} & \textbf{79.48} & \textbf{87.14} & \textbf{87.36} \\
\bottomrule
\end{tabular}
\end{table*}

\subsection{Baselines and Evaluation Metrics}

To comprehensively evaluate the effectiveness of the proposed \textbf{SPICE} framework, we compare it against a diverse set of multimodal learning baselines, including classical fusion strategies, modality rebalancing approaches, and curriculum-based methods.

The \textbf{classical fusion-based baselines} include \emph{Concat}, \emph{Affine}~\cite{perez2018film}, \emph{Channel}, \emph{ML-LSTM}~\cite{nie2021multimodal}, \emph{Sum}, \emph{Weight}, and \emph{ETMC}~\cite{han2023trusted}. These methods represent traditional multimodal fusion paradigms that combine modality-specific representations through concatenation, weighted aggregation, or learned fusion operators.

The \textbf{modality balancing and optimization-based baselines} include \emph{MSES}~\cite{fujimori2019modality}, \emph{OGR-GB}~\cite{wang2019WhatMT}, \emph{DOMFN}~\cite{yang2022domfn}, \emph{OGM}~\cite{peng2022ogm}, \emph{MSLR}~\cite{yao2022modality}, \emph{AGM}~\cite{li2023agm}, \emph{PMR}~\cite{fan2023pmr}, \emph{ReconBoost}~\cite{hua2024reconboost}, \emph{MMPareto}~\cite{wei2024mmpareto}, \emph{SMV}~\cite{wei2024smv}, \emph{MLA}~\cite{zhang2024mla}, and \emph{AMSS}~\cite{yang2025amss}. These baselines focus on addressing modality imbalance through optimization control, adaptive fusion, sample-level valuation, and representation rebalancing.

We further compare against recent \textbf{curriculum-based methods}, including \emph{BSS-H}~\cite{guan2025balanceaware} and \emph{BSS-L}~\cite{guan2025balanceaware}, which develop training schedules based on loss and similarity measures. Following~\cite{peng2022ogm,hua2024reconboost}, we report \textbf{Accuracy (ACC)}, \textbf{Mean Average Precision (mAP)}, and \textbf{Macro F1-score (Mac-F1)} as evaluation metrics. Accuracy measures the proportion of correctly classified samples, mAP captures the average precision across all samples, while Mac-F1 computes the average of F1 scores across all categories.

\subsection{Results and Analysis}

\subsubsection{Overall Results}

Table~\ref{tab:main_results} presents a quantitative comparison of the proposed \textbf{SPICE} framework against various fusion-based, optimization-based, and sequence sampling baselines across three benchmark datasets: CREMA-D, Kinetics-Sounds, and NVGesture. The benchmark results of baselines are adopted from the BSS paper~\cite{guan2025balanceaware}, following their pre-processing and evaluation protocol.

Overall, the proposed \textbf{SPICE-E} consistently achieves the best performance across all evaluated benchmarks, demonstrating the effectiveness of dynamically modeling multimodal sample difficulty using PID-inspired interaction scores. 

Please note that the “-” in Table~\ref{tab:main_results} indicates that the corresponding methods do not apply to those datasets.

\subsubsection{Results on Bimodal Datasets}
On \textbf{CREMA-D}, SPICE-E achieves the highest performance with an accuracy of \textbf{83.06\%} and mAP of \textbf{89.07\%}, outperforming the strongest prior baseline BSS-L by \(+0.26\%\) in ACC and \(+0.46\%\) in mAP. While the numerical gains may appear modest, they are significant given the already saturated performance regime of recent multimodal rebalancing approaches.

A similar trend is observed on \textbf{Kinetics-Sounds}, where SPICE-E achieves \textbf{73.99\%} ACC and \textbf{79.48\%} mAP, again surpassing BSS-L and all other baselines. The improvement over BSS-L is particularly notable in terms of mAP, suggesting that the proposed curriculum better preserves ranking consistency across samples and categories.

\subsubsection{Results on Trimodal Dataset}
On the more challenging \textbf{NVGesture} benchmark, which involves three modalities, SPICE-E obtains the best overall performance with \textbf{87.14\%} ACC and \textbf{87.36\%} F1-score. This improvement over BSS-L (\(86.72\%\) ACC, \(87.04\%\) F1) highlights the scalability of the proposed PID-guided curriculum beyond bimodal settings.

\subsubsection{Results on Large-Scale Dataset}

\begin{table}[t]
\caption{Performance comparison on the VGGSound dataset. Best results are shown in \textbf{bold}.}
\label{tab:vgg_results}
\small
\begin{tabular}{lcc}
\toprule
\textbf{Method} & \textbf{ACC (\%)} & \textbf{mAP (\%)} \\
\midrule
OGM         & 48.29 & 49.78 \\
AGM         & 47.11 & 51.98 \\
ReconBoost  & 50.97 & 53.87 \\
MMPareto    & 51.25 & 54.73 \\
SMV         & 50.31 & 53.62 \\
MLA         & 51.65 & 54.73 \\
BSS-H       & 51.61 & 55.68 \\
BSS-L       & 52.80 & 56.61 \\
\midrule
SPICE-E     & \textbf{54.98} & \textbf{56.63} \\
\bottomrule
\end{tabular}
\end{table}

To evaluate the proposed framework's scalability, we conducted experiments on the large-scale \textbf{VGGSound} benchmark. As shown in Table~\ref{tab:vgg_results}, BSS-L offers the best baseline among prior multimodal learning methods.

By extending SPICE to a larger, diverse benchmark, we explored the effectiveness of the PID-guided dynamic curricula beyond medium-scale datasets like CREMA-D, Kinetics-Sounds, and NVGesture. The results show that SPICE-E maintains strong performance in large-scale scenarios, indicating that the dynamic approach to sample difficulty ordering generalizes well across different dataset scales and complexities.

\subsubsection{Sample Binning vs. Ordering}
\label{sec:compairng-SPICES-SPICEE}
Comparing \textbf{SPICE-S} and \textbf{SPICE-E} highlights the importance of the type of dynamic scheduling. \textbf{SPICE-S} divides samples into different bins, and uses different bin combinations in different training stages. Only a subset of bins are used in the early stages. \textbf{SPICE-E}, on the other hand, utilizes all samples in all the training stages, but orders them using different strategies based on PID scores in different training stages. 

While SPICE-S achieves strong performance,  SPICE-E consistently outperforms it. This suggests that using all samples with continuously updated ordering is more effective than stage-wise sample selection. The gains are particularly pronounced on CREMA-D, where SPICE-E improves over SPICE-S by \(+2.15\%\) for ACC and \(+2.31\%\) for mAP. 

\begin{figure}
  \centering
  \includegraphics[width=0.98\linewidth]{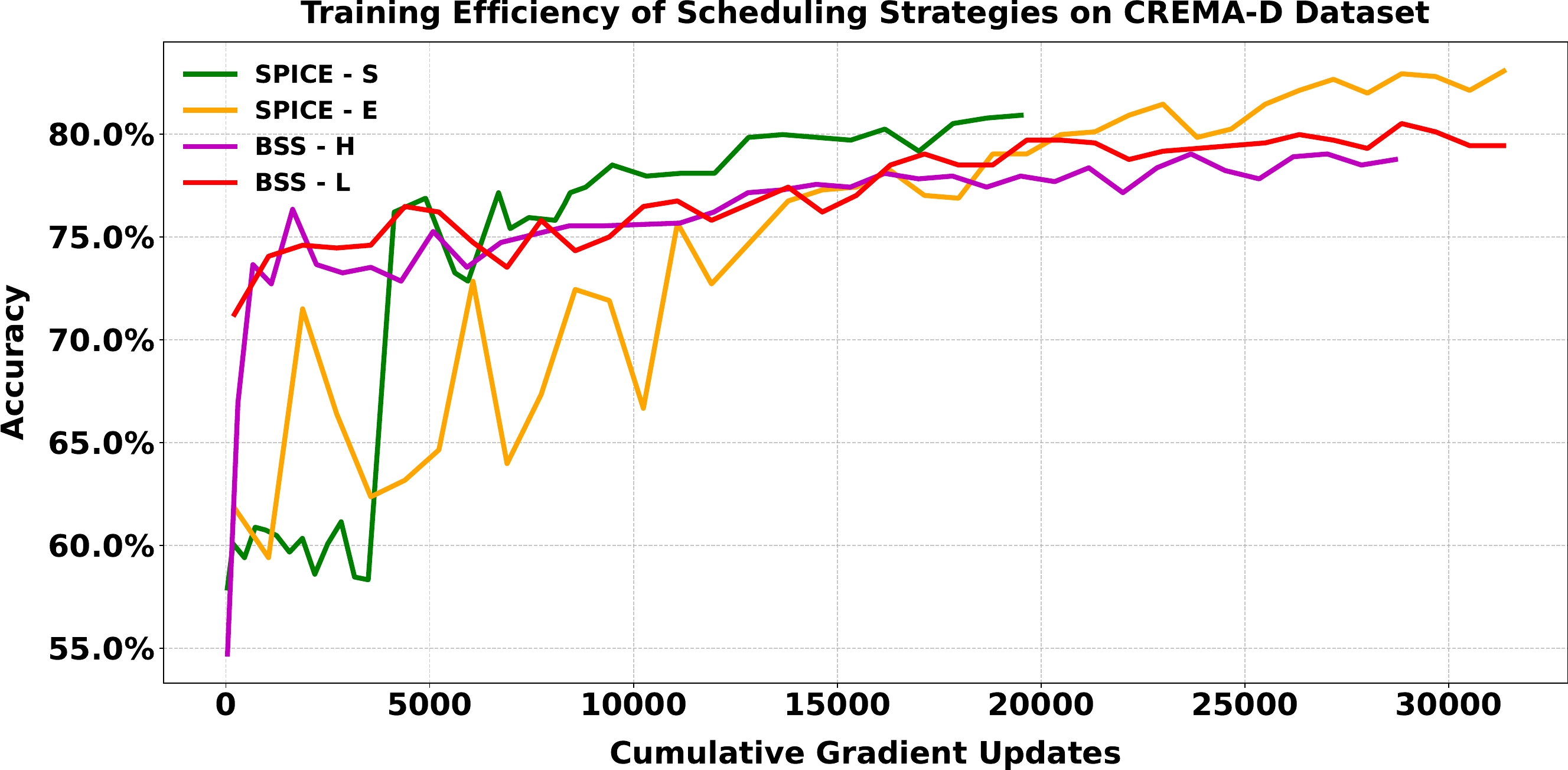}
  \caption{Cost comparison in terms of gradient updates for SPICE-S, SPICE-E, BSS-H, and BSS-L.}
  \Description{Comparing the total number of gradient updates needed to reach a particular accuracy.}
  \label{fig:cost-compare}
\end{figure}

One potential advantage of SPICE-S is its training efficiency at the early stages, where only a subset of samples is used. 
Figure \ref{fig:cost-compare} illustrates that SPICE-S achieves high accuracy with fewer gradient updates than SPICE-E and the BSS baseline. This indicates faster convergence and greater training efficiency in the early stages. SPICE-E gradually catches up and surpasses SPICE-S in the final accuracy, demonstrating its long-term performance gain with more gradient updates. Overall, SPICE-S is clearly a cost-effective method for training the multimodal network, as it offers comparable performance (2.15\% decrease in final accuracy) with nearly one-third fewer gradient updates.

\subsubsection{Discussion on Prior Sequence Sampling Methods}

A particularly important comparison can be made with the balance-aware sequence sampling baselines \textbf{BSS-H} and \textbf{BSS-L}, which also employ sample scheduling strategies for dynamic curriculum learning.

Unlike BSS methods, which rely on a multi-perspective balance measurer combining correlation and information criteria, SPICE dynamically updates the curriculum throughout training using model-driven estimates of redundancy, unique information, and synergy. Specifically, BSS quantifies sample difficulty via a balance score that integrates prediction consistency across modalities (correlation criterion) with label-dependent training loss (information criterion), and then constructs a training sequence that prioritizes more “balanced” samples. While its heuristic scheduler follows a predefined pacing function over a fixed ranking, its learning-based variant introduces probabilistic reweighting of samples with periodic updates. Although effective, this design still relies on proxy measures of agreement and correctness, and the curriculum is only indirectly influenced by the evolving model. In contrast, SPICE avoids such proxy measures and instead leverages endogenous unimodal and multimodal predictions to directly estimate partial information components.  The consistent gains of SPICE-E over BSS-L across all datasets validate our central hypothesis that \emph{multimodal sample difficulty evolves with the training progress and can be dynamically estimated from endogenous unimodal and multimodal predictions through the lens of PID}. Consequently, SPICE adapts the sampling strategy (Sample Binning or Ordering discussed in Sec.~\ref{sec:compairng-SPICES-SPICEE}) in a more principled and fine-grained manner as modality representations and the fusion model mature.

\subsubsection{Why Does SPICE Work?}

The superior performance of SPICE can be attributed to its ability to explicitly disentangle different forms of multimodal information during training.

In the early stages, the curriculum prioritizes  \textbf{redundant} information, allowing the model to first learn stable, low-level shared representations across modalities.

As training progresses, the curriculum gradually shifts its focus toward \textbf{unique} and \textbf{synergistic} samples, which require modality-specific reasoning and higher-order interaction modeling.

This easy-to-complex progression is realized either through stage-wise sample inclusion or continuous reordering, and aligns naturally with the evolution of learned representations, providing a more meaningful notion of difficulty.

\section{Curriculum based on Annotator Responses}

To better understand the role of curriculum construction, we conduct a series of ablation studies based on \textbf{annotator-derived difficulty signals} on the CREMA-D dataset~\cite{cao2014cremad}.  In CREMA-D, each clip has an intended emotion (the ground-truth label) and receives perceived emotion labels from crowdsourced human annotators. Since multiple annotators rate the same clip, their responses can vary, creating an annotation distribution that reflects different perceptions of the emotion. Annotator-derived responses are the labels assigned by human raters who evaluate each clip.

\subsection{Static Curriculum from Annotator Responses}

In the first setting, we construct a \textbf{static curriculum} based solely on annotator responses and agreement scores. Specifically, the model predictions and confidence scores used in SPICE are replaced by annotator-derived responses. The curriculum is fully predefined before training begins and remains unchanged throughout the training. To ensure fair comparison, training follows the same sampling strategy and uses the same number of epochs as SPICE-E. This study helps to isolate the contribution of \emph{human-derived static difficulty priors} from the proposed model-adaptive curriculum.

\subsection{Hybrid Adaptive Curriculum}

We further investigate a \textbf{hybrid adaptive PID curriculum} that progressively transitions from PID-based sample ordering based on the exogenous annotator responses to unimodal and multimodal predictions generated by the model under training. Hybrid scores at epoch \(t\) are calculated as
\begin{equation}
D_t^{(c)} =
(1-\alpha_t)D_{\text{annotator}}^{(c)}
+
\alpha_t D_{\text{model}}^{(c)},
\quad
c \in \{R,U,S\}, 
\end{equation}
where \(D_{\text{annotator}}^{(c)}\) denotes the static PID score derived from annotator responses and \(D_{\text{model}}^{(c)}\) denotes the dynamic PID score estimated from the current model state. We evaluate two weighting functions: 1) $\alpha_t = \frac{t}{T}$, which provides a linear convergence  towards model predictions; 2) $\alpha_t =
\frac{1-\cos\left(\pi t / T\right)}{2}$, which enables a smoother non-linear (Cosine) shift from annotator priors to model predictions.

These experiments are designed to analyze how curriculum difficulty should evolve and whether a hybrid human-model curriculum provides additional gains over fully model-driven SPICE.

\subsection{Analysis of Annotator-Guided Curriculum Variants}

\begin{table}[t]
\caption{PID curricula based on annotator responses for CREMA-D Dataset. Best results are shown in \textbf{bold}.}
\label{tab:ablation_annotator}
\small
\begin{tabular}{lcc}
\toprule
\textbf{Method} & \textbf{ACC (\%)} & \textbf{mAP (\%)} \\
\midrule
Complete Static Curriculum & 76.75 & 83.59 \\
Hybrid Linear Adaptive Curriculum & 71.91 & 78.88 \\
Hybrid Non-Linear Adaptive Curriculum & 72.18 & 79.30 \\
\midrule
SPICE-S & 80.91 & 86.76 \\
\textbf{SPICE-E} & \textbf{83.06} & \textbf{89.07} \\
\bottomrule
\end{tabular}
\end{table}

Table~\ref{tab:ablation_annotator} presents the ablation results for different curriculum strategies derived from annotator responses.

Fully model-driven dynamic curriculum learning consistently outperforms annotator-informed curricula. The proposed \textbf{SPICE-E} achieves the highest performance, significantly surpassing all annotator-based alternatives. Among the annotator-guided variants, the \textbf{complete static curriculum} achieves the strongest performance (\(76.75\%\) ACC, \(83.59\%\) mAP).

The significant performance gap between the static curriculum and SPICE-E (\(+6.31\%\) ACC and \(+5.48\%\) mAP) strongly suggests that \emph{human-derived difficulty alone is insufficient for optimal multimodal learning}. While annotators may capture perceived ambiguity or disagreement, these signals remain fixed and do not adapt to the model's evolving representational capacity.

Interestingly, the  \textbf{hybrid adaptive curricula} perform notably worse than the fully static variant. This observation suggests that naively interpolating between annotator priors and model-driven sample ordering may introduce instability in the curriculum transition. During the intermediate training stages, the annotator-derived and model-derived signals may provide conflicting notions of difficulty, leading to suboptimal sampling and poor convergence.

The results further highlight that sample difficulty in multimodal interaction learning is not purely an intrinsic property of the data, but rather a \textbf{dynamic and model-dependent quantity}. As training progresses, samples that are initially difficult may become easier once shared representations mature, while higher-order synergistic interactions may only emerge in later stages.

This directly supports the central motivation of SPICE: \textbf{sample ordering should be dynamically adjusted based on the current model state rather than being fixed a priori}.

\section{Conclusion and Future Work}

We propose \textbf{SPICE}, a PID-guided progressive curriculum framework for multimodal interaction learning that models sample difficulty as a \emph{dynamic and evolving quantity}. By decomposing multimodal interactions into \textbf{redundant}, \textbf{unique}, and \textbf{synergistic} components, SPICE enables interpretable, adaptive curriculum development. Extensive experiments show that SPICE outperforms conventional fusion, modality rebalancing, and balance-aware sequence sampling methods, supporting our hypothesis that sample ordering should be updated continuously throughout training based on sample difficulty. This curriculum applies to systems such as social robots and adaptive human-computer interfaces, where modality reliability fluctuates. Because the proposed scores rely solely on model predictions, SPICE can adapt online to changing conditions. Future work will investigate using annotator priors as uncertainty-aware regularizers rather than direct curriculum signals and will extend SPICE to large multimodal foundation models and sequential interaction settings.

\bibliographystyle{ACM-Reference-Format}
\bibliography{spice-refs}

\end{document}